\documentclass[runningheads]{llncs}
\usepackage[T1]{fontenc}
\usepackage{graphicx,verbatim}
\usepackage{array}
\usepackage{xcolor}
\usepackage{amsmath,amssymb,amsfonts}
\usepackage{epstopdf}
\usepackage{subfigure}
\usepackage{hyperref}
\usepackage{marvosym}
\begin{document}
\title{AnomExpert: Identifying and Selecting Anatomical Planes for Prenatal Ultrasound Anomaly Diagnosis}
\titlerunning{AnomExpert}
%

\author{
Jian Wang\inst{1,2,3}\thanks{Jian Wang and Yang Yang contributed equally to this work.} \and
Yang Yang\inst{4}\textsuperscript{$\star$} \and
Ziheng Pan\inst{5} \and
Xiliang Zhu\inst{5} \and
Yuhan Zhang\inst{4} \and
Yanfeng Zhou\inst{1,3} \and
Dong Ni\inst{1,3,5}\textsuperscript{(\Letter)}
}

\authorrunning{J. Wang et al.}

\institute{
Medical Ultrasound Image Computing (MUSIC) Lab, School of Artificial Intelligence, Shenzhen University, Shenzhen 518060, China
\and
College of Computer Science and Software Engineering, Shenzhen University, Shenzhen 518060, China
\and
National Engineering Laboratory for Big Data System Computing Technology, Shenzhen University, Shenzhen 518060, China
\and
School of Biomedical Engineering, Medical School, Shenzhen University, Shenzhen 518037, China
\and
School of Biomedical Engineering and Informatics, Nanjing Medical University, Nanjing 211166, China\\
\email{nidong@szu.edu.cn}
}
\maketitle              

\begin{abstract}
Life-limiting congenital anomalies require accurate prenatal diagnosis for appropriate clinical decision-making. 
Prenatal ultrasound (US) examinations involve multiple anatomical planes, and diagnosis depends on identifying anatomical planes and selecting diagnostically relevant planes for each anomaly. 
Existing automated methods either rely on plane-level annotations or aggregate heterogeneous images without explicitly modeling these diagnostic capabilities.
We propose \textbf{AnomExpert}, a prototype-driven framework for prenatal US anomaly diagnosis using only case-level supervision. 
AnomExpert introduces learnable plane prototypes to organize unordered images into latent representations corresponding to anatomical planes without requiring plane annotations. 
A disease-aware sparse selection mechanism further selects diagnostically relevant planes for each anomaly.
Experiments on a multi-center dataset of 3,654 cases show that AnomExpert consistently outperforms nine representative multi-instance learning methods. 
Using a ViT-small backbone, it achieves 86.9\% accuracy and 84.2\% F1-score while maintaining parameter efficiency. 
These findings indicate that modeling anatomical plane identification and disease-specific plane selection improves weakly supervised multi-plane prenatal US anomaly classification.
The code is available at \href{https://github.com/TIanCat/AnomExpert}{https://github.com/TIanCat/AnomExpert}.


\end{abstract}
\section{Introduction}

\begin{figure}[b]
\centering
\includegraphics[scale=0.3]{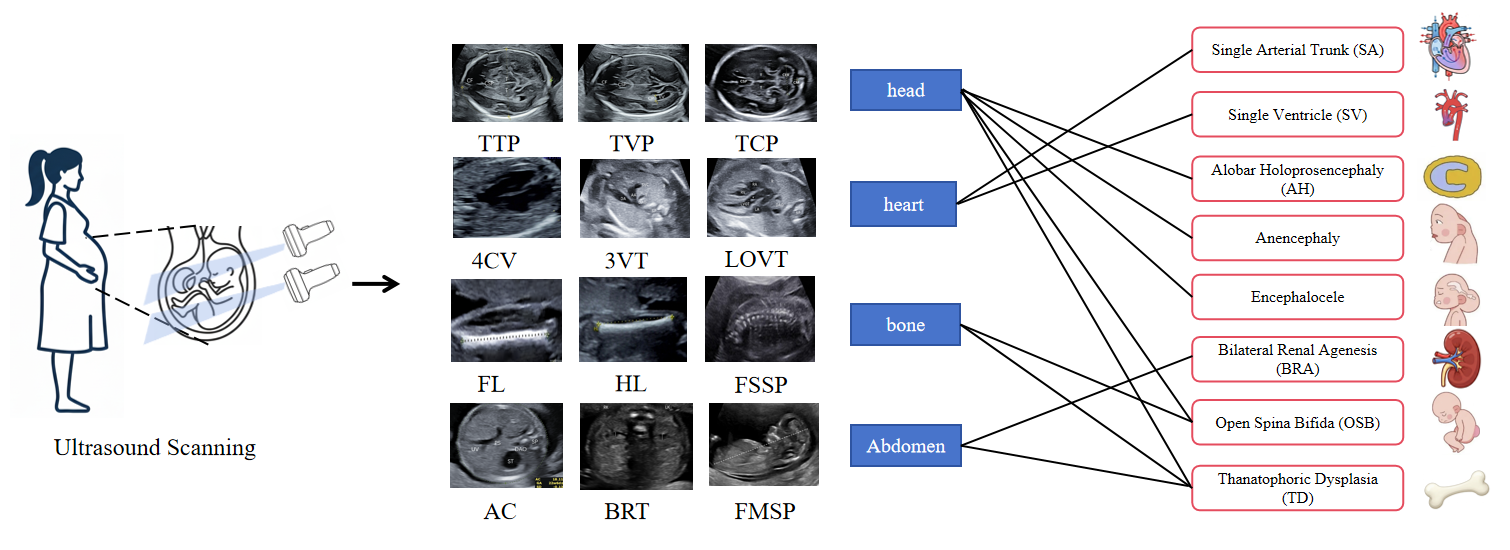}
\caption{Prenatal Ultrasound Anomaly Diagnosis}
\label{figure1}
\end{figure}

Life-limiting congenital anomalies (LLCAs) are severe fetal conditions associated with high mortality~\cite{breeze2013antenatal}. 
Different anomalies require distinct management strategies. 
For example, severe open spina bifida may warrant prenatal fetal repair~\cite{adzick2011randomized}, complex congenital heart defects often require staged surgery after birth~\cite{ohye2016current}, and anencephaly is typically managed with palliative care or pregnancy termination~\cite{cook2008prenatal}. 
Therefore, accurate prenatal differentiation among anomaly types is critical for appropriate clinical decision-making. 
Ultrasound (US) is the primary modality for screening and diagnosis~\cite{salomon2022isuog}.
In clinical practice, a typical prenatal US examination involves acquiring images from multiple anatomical planes. 
\emph{Accurate diagnosis requires identifying the anatomical plane represented by each image and determining which planes provide diagnostically relevant evidence for a specific anomaly} (Fig.~\ref{figure1}). 
However, this diagnostic process is time-consuming and cognitively demanding, as different anomalies rely on distinct subsets of anatomical planes. 
Moreover, US interpretation is inherently subjective, leading to variability in diagnostic consistency~\cite{sarris2012intra}.  
These challenges motivate automated methods for multi-plane prenatal diagnosis.

Developing automated systems that capture these diagnostic capabilities remains challenging.
Existing approaches often follow staged pipelines, such as standard plane detection followed by anomaly classification~\cite{guo2022fetal,arnaout2021ensemble}.
While effective under curated conditions, these methods require fine-grained image-level annotations, which are costly and difficult to obtain.
Moreover, they assume that standard anatomical planes can be reliably localized, an assumption often violated in clinical practice due to anatomical variability and abnormal presentations~\cite{coronado2023automatic,huang2023detecting}.
Multi-Instance Learning (MIL) provides a natural alternative by learning from unordered image sets using case-level supervision~\cite{barbosa2024multiple}. 
However, most MIL methods aggregate image features directly through attention or ranking mechanisms without explicitly modeling anatomical plane identity or disease-specific plane selection~\cite{ilse2018attention,ilse2020deep,shao2021transmil,li2021dual,javed2022additive,zhang2024mamba2mil,shiku2025ordinal,liang2025medical}. 
As a result, heterogeneous anatomical planes are mixed within a unified representation, limiting the model’s ability to capture plane-level semantics and disease-specific relevance.

Motivated by the fundamental capabilities required for multi-plane diagnosis, we propose \textbf{AnomExpert}, a prototype-driven framework for prenatal US anomaly diagnosis using case-level supervision. 
Instead of replicating a staged pipeline, AnomExpert explicitly models two key capabilities: anatomical plane identification and disease-specific plane selection. 
Specifically, learnable plane prototypes organize unordered images into latent representations corresponding to anatomical planes without requiring plane annotations. 
A disease-aware sparse selection mechanism then adaptively identifies diagnostically relevant planes for each anomaly. 
This design enables the model to differentiate anatomical planes under case-level supervision and to selectively integrate disease-relevant planes for diagnosis.
Our contributions are threefold: 
(1) a prototype-driven framework for weakly supervised anatomical plane identification without plane annotations; 
(2) a disease-aware sparse aggregation mechanism for adaptive diagnostic plane selection; and 
(3) extensive experiments demonstrating consistent improvements over representative MIL methods for prenatal US anomaly classification.

\section{Methodology}
\begin{figure}[htpb]
\centering
\includegraphics[scale=0.355]{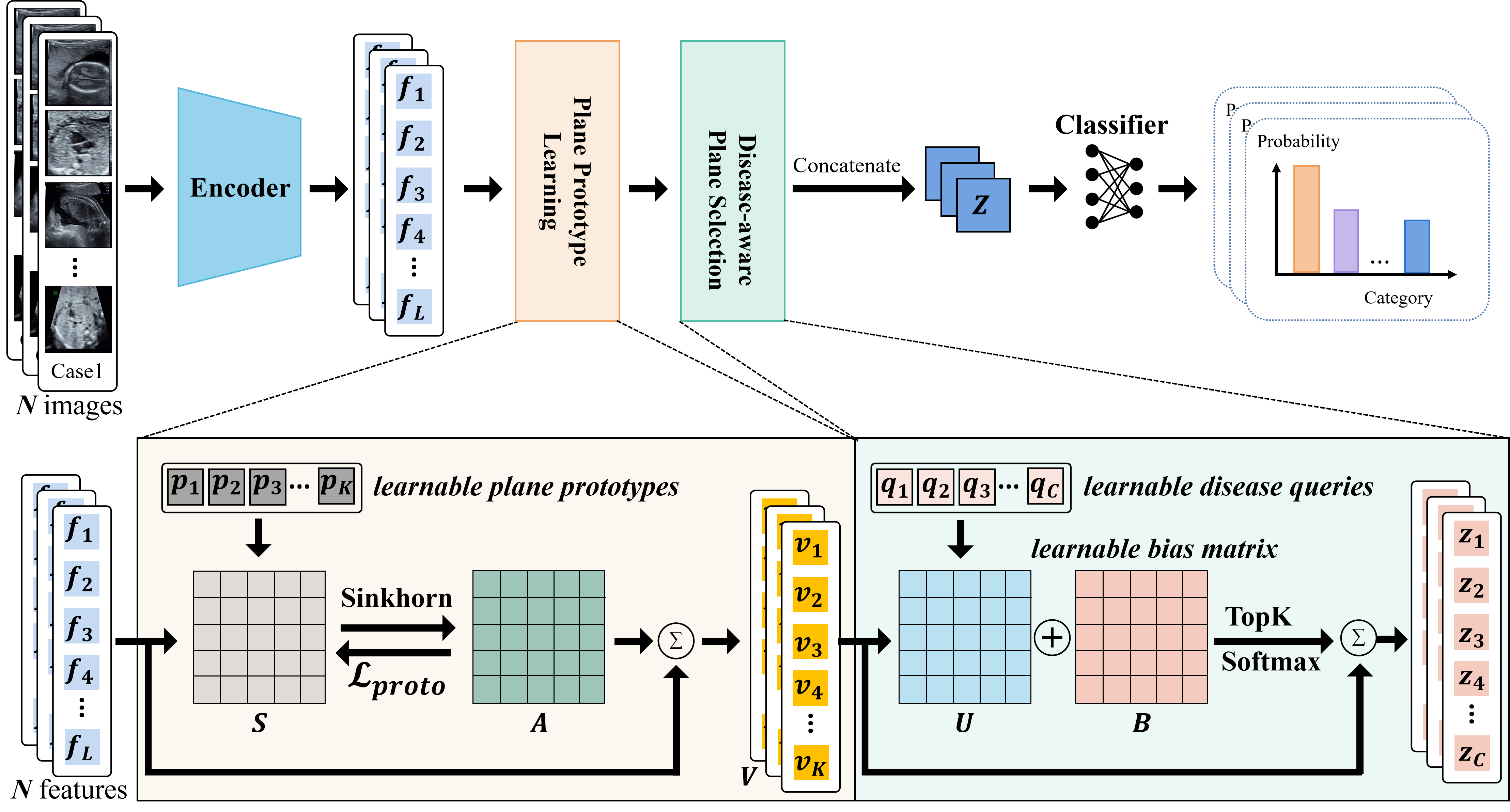}
\caption{ Overview of the proposed AnomExpert }
\label{figure2}
\end{figure}

As illustrated in Fig.~\ref{figure2}, AnomExpert models two key capabilities of prenatal US diagnosis: anatomical plane identification and disease-specific plane selection. 
The framework first organizes unordered images into plane-level representations using learnable plane prototypes.
These plane representations serve as candidate anatomical planes within each case.
A disease-aware selection mechanism then identifies diagnostically relevant planes for anomaly prediction.
The entire framework is trained end-to-end using only case-level labels.

\subsection{Plane Prototype Learning}
\label{ppl}
A key challenge in multi-plane US diagnosis is identifying anatomical planes without explicit plane annotations. 
To approximate anatomical plane identification under case-level supervision, we introduce $K$ learnable plane prototypes.
Each prototype is intended to capture a distinct anatomical plane pattern. 
During training, image features are softly assigned to these prototypes, encouraging images with similar anatomical content to cluster around the same prototype.

\textbf{Image embedding:}
Given an input image $x_i$, we extract a feature representation using a backbone network $\phi(\cdot)$ followed by a projection head $g(\cdot)$:
\begin{equation}
f_i = g(\phi(x_i)) \in \mathbb{R}^{256}.
\end{equation}

\textbf{Prototype assignment:}
We maintain a set of learnable plane prototypes $P=\{p_k\in\mathbb{R}^{256}\}_{k=1}^{K}$. 
For a mini-batch containing $N$ images (summed across multiple cases), we compute cosine similarities between all image features and prototypes to form a score matrix $S\in\mathbb{R}^{N\times K}$.  
Soft assignments are obtained using Sinkhorn-balanced normalization~\cite{cuturi2013sinkhorn}:
\begin{equation}
A=\mathrm{Sinkhorn}\!\left(\exp(S/\epsilon_s)\right) \in\mathbb{R}^{N\times K},
\end{equation}
where $\epsilon_s=0.05$ controls assignment sharpness. 
Sinkhorn balancing encourages diverse prototype utilization and helps prevent prototype collapse.

\textbf{Prototype loss:}
To encourage consistent image-to-prototype assignments, we optimize a prototype assignment loss:
\begin{equation}
\mathcal{L}_{proto}
=
-\frac{1}{N}\sum_{i=1}^{N}\sum_{k=1}^{K}
A_{i,k}\log \mathrm{softmax}(S_{i,:}/\tau)_k,
\label{eq:proto}
\end{equation}
where $\tau$ is a learnable temperature parameter. 
Through this learning process, images with similar anatomical characteristics are grouped under the same prototype, enabling anatomical plane identification without plane labels.

\textbf{Plane representation:}
Given a case with $L$ images, we aggregate image features into plane representations using the corresponding assignment weights:
\begin{equation}
v_k=
\frac{\sum_{i=1}^{L} A_{i,k} f_i}{\sum_{i=1}^{L} A_{i,k}+\delta},
\quad k=1,\dots,K,
\end{equation}
where $\delta$ is a small constant for numerical stability. 
This yields a structured plane representation set $V=\{v_k\in \mathbb{R}^{256}\}_{k=1}^{K}$ for the current case, which serves as the basis for downstream disease-specific plane selection.

\subsection{Disease-aware Plane Selection}
\label{dps}
Unlike conventional MIL approaches that aggregate features from all images, our method performs selection at the plane-level.
To this end, we introduce learnable disease queries to estimate the relevance of each anatomical plane to each disease.

\textbf{Disease queries:}
We define a set of learnable disease queries:
\begin{equation}
Q = \{q_c \in \mathbb{R}^{256}\}_{c=1}^{C},
\end{equation}
where $C$ is the number of categories.

\textbf{Relevance estimation:}
Given the structured plane representations $V=\{v_k\in \mathbb{R}^{256}\}_{k=1}^{K}$ obtained in Sec.~\ref{ppl}, we compute cosine similarity scores between each disease query and all plane representations:
\begin{equation}
u_{c,k} = \mathrm{sim}(q_c, v_k).
\end{equation}
To capture disease-specific preferences over anatomical planes, we introduce a learnable bias matrix $B \in \mathbb{R}^{C \times K}$, which models prior associations between diseases and planes. The final relevance scores are obtained as:
\begin{equation}
\tilde{u}_{c,k} = u_{c,k} + B_{c,k}.
\end{equation}

\textbf{Sparse plane selection and aggregation:}
For each disease $c$, we retain the top-$k$ most relevant planes:
\begin{equation}
\Omega_c = \mathrm{TopK}(\tilde{u}_{c,:}),
\end{equation}
where $\Omega_c$ denotes the selected plane indices.
We then compute disease-specific representations by aggregating the selected planes using normalized weights:
\begin{equation}
z_c =
\sum_{k \in \Omega_c}
\alpha_{c,k} v_k,
\qquad
\alpha_{c,\Omega_c}
=
\mathrm{softmax}(\tilde{u}_{c,\Omega_c}/T),
\end{equation}
where $T=0.07$ is a temperature constant controlling the sharpness of selection.
This mechanism ensures that anomaly prediction is based on a subset of diagnostically relevant anatomical planes rather than all available images.

\textbf{Case-level prediction:}
The disease-specific representations are concatenated and fed into a classifier to obtain the final prediction:
\begin{equation}
\hat{y} = \mathrm{Classifier}([z_1,\dots,z_C]).
\end{equation}

\textbf{Training objective:}
The model is trained end-to-end using case-level category label $y$. 
We use cross-entropy loss for anomaly classification:
\begin{equation}
\mathcal{L}_{cls}
=
\mathrm{CE}(\hat{y}, y),
\end{equation}
and jointly optimize the prototype assignment loss defined in Equation~\ref{eq:proto}:
\begin{equation}
\mathcal{L}
=
\mathcal{L}_{cls}
+
\lambda \mathcal{L}_{proto},
\end{equation}
where $\lambda$ balances plane prototype learning and disease classification.

\section{Experiments and Results}

\subsection{Datasets and Experimental Setup}

\begin{table*}[t]
\caption{Performance comparison of different methods on the test set (mean ± std over five runs). Param. denotes model parameters (M), and all metrics are in percentage (\%). Best and second-best results are shown in blue and underlined, respectively.}
\label{tab1}
\centering
\begin{tabular}{l|c>{\centering\arraybackslash}p{1.65cm} >{\centering\arraybackslash}p{1.65cm} >{\centering\arraybackslash}p{1.65cm} >{\centering\arraybackslash}p{1.65cm} >{\centering\arraybackslash}p{1.65cm}}
\hline
\textbf{Methods} & \textbf{Param.} & \textbf{Accuracy} & \textbf{Precision} & \textbf{Recall} & \textbf{F1-score}  & \textbf{AUC}  \\
\hline
AttentionMIL  & 11.31  & 81.7$\pm$1.08 & 80.0$\pm$1.79  & 78.2$\pm$1.24  &78.4$\pm$1.39    &  96.4$\pm$0.47          \\
MeanNetMIL & 23.53  & 82.2$\pm$0.80 & 80.9$\pm$1.00 & 79.3$\pm$1.50 & 79.3$\pm$1.40 & 96.7$\pm$0.30 \\
MaxNetMIL  & 21.29  & \underline{85.5$\pm$1.06} & \underline{84.0$\pm$1.52} & \underline{81.5$\pm$1.47} & \underline{82.1$\pm$1.59} & 96.7$\pm$0.32 \\
TransMIL   & 30.91  & 83.5$\pm$1.30 & 82.2$\pm$2.00 & 81.0$\pm$2.01 & 81.2$\pm$2.00 & 96.9$\pm$0.50 \\
DSMIL      & 23.97  & 78.1$\pm$2.54 & 77.2$\pm$2.40 & 75.8$\pm$2.84 & 75.0$\pm$2.82 & 95.8$\pm$0.60 \\
AdditiveMIL  & 11.97  &77.9$\pm$0.91   &74.7$\pm$1.52  &73.7$\pm$1.31  & 74.0$\pm$1.40      &     95.8$\pm$0.34        \\
Mamba2MIL  & 39.43  & 82.3$\pm$4.49  & 82.1$\pm$3.78  & 81.8$\pm$3.99  & 81.2$\pm$4.05   & 97.1$\pm$0.97   \\
SAMIL  & 27.72    &  84.1$\pm$0.52    &82.2$\pm$1.49   & 80.0$\pm$0.83  & 80.3$\pm$1.05      &  96.4$\pm$0.41           \\
AAcls-MIL  & 19.07  & 83.6$\pm$1.18 & 82.3$\pm$1.28 & 79.8$\pm$1.03 & 80.3$\pm$0.86 & 97.3$\pm$0.31 \\

\textbf{Anom.}(ViT-t) & 5.60 & 83.8$\pm$0.96 & 81.8$\pm$0.93 & 80.4$\pm$1.39 & 80.6$\pm$1.24 & \underline{97.3$\pm$0.23} \\
\textbf{Anom.}(ViT-s) & 21.80  & \textcolor{blue}{86.9$\pm$0.72} & \textcolor{blue}{84.9$\pm$1.12} & \textcolor{blue}{83.9$\pm$0.66} & \textcolor{blue}{84.2$\pm$0.69} & \textcolor{blue}{97.9$\pm$0.20} \\
\hline
\end{tabular}
\end{table*}

\subsubsection{Datasets}
In this study, we collected a multi-center prenatal US dataset from 24 medical centers with approval from the institutional review board; informed consent was waived due to the retrospective study design. 
The dataset comprises 3,654 cases with 61,460 images, averaging 17 images per case (range: 5–94, std$=$13).
Each case was assigned a single primary diagnosis. 
All diagnoses were confirmed by experienced fetal US specialists according to established clinical criteria.
The dataset includes eight types of LLCAs and normal controls: Single Arterial Trunk (SA, n$=$399), Single Ventricle (SV, n$=$341), Alobar Holoprosencephaly (AH, n$=$478), Anencephaly (n$=$643), Encephalocele (n$=$180), Bilateral Renal Agenesis (BRA, n$=$335), Open Spina Bifida (OSB, n$=$348), Thanatophoric Dysplasia (TD, n$=$410), and Normal (n=520). 
Gestational ages ranged from 10 to 39 weeks (mean ± std: 20.7 ± 12.0 weeks). 
The dataset was randomly partitioned at the case level into training (n$=$2,558), validation (n$=$366), and testing (n$=$730) sets to prevent data leakage.
The splits were stratified to preserve class distribution, and gestational age distributions were comparable across subsets.

\subsubsection{Implementation Details}
We evaluated our method using two ImageNet-pretrained backbone architectures, ViT-tiny (ViT-t) and ViT-small (ViT-s), to assess robustness across different model capacities.
The number of plane prototypes was set to $K=30$, and the disease-aware selection retained the top-$k=4$ planes for each category. 
The weight of the prototype assignment loss was fixed at $\lambda=0.1$. 
Models were trained for 60 epochs using Adam (initial learning rate $1\times10^{-4}$, weight decay $1\times10^{-5}$), and the checkpoint with the best validation performance was used for testing.
We used cosine annealing for learning rate decay and a batch size of 8 (case-level). 
Input images were randomly cropped and resized to $224\times224$. 
Additional data augmentations included horizontal flipping, color jittering, affine transformations, and random grayscale conversion.

\subsubsection{Evaluation Protocol}
We compared our method with nine representative multi-instance learning (MIL) approaches, including AttentionMIL~\cite{ilse2018attention}, MeanNetMIL~\cite{ilse2020deep}, MaxNetMIL~\cite{ilse2020deep}, TransMIL~\cite{shao2021transmil}, DSMIL~\cite{li2021dual}, AdditiveMIL~\cite{javed2022additive}, Mamb-a2MIL~\cite{zhang2024mamba2mil}, SAMIL~\cite{shiku2025ordinal}, and AAcls-MIL~\cite{liang2025medical}. 
All competing methods were implemented using their official open-source code and evaluated under the same data split.
Performance was evaluated using Accuracy, macro-averaged Precision, Recall, F1-score, and area under the ROC curve (AUC). 
Each method was trained and evaluated over five independent runs with different random seeds. We report the mean and standard deviation across runs.

\subsection{Results}

\begin{figure}[t]
\centering
\includegraphics[scale=0.33]{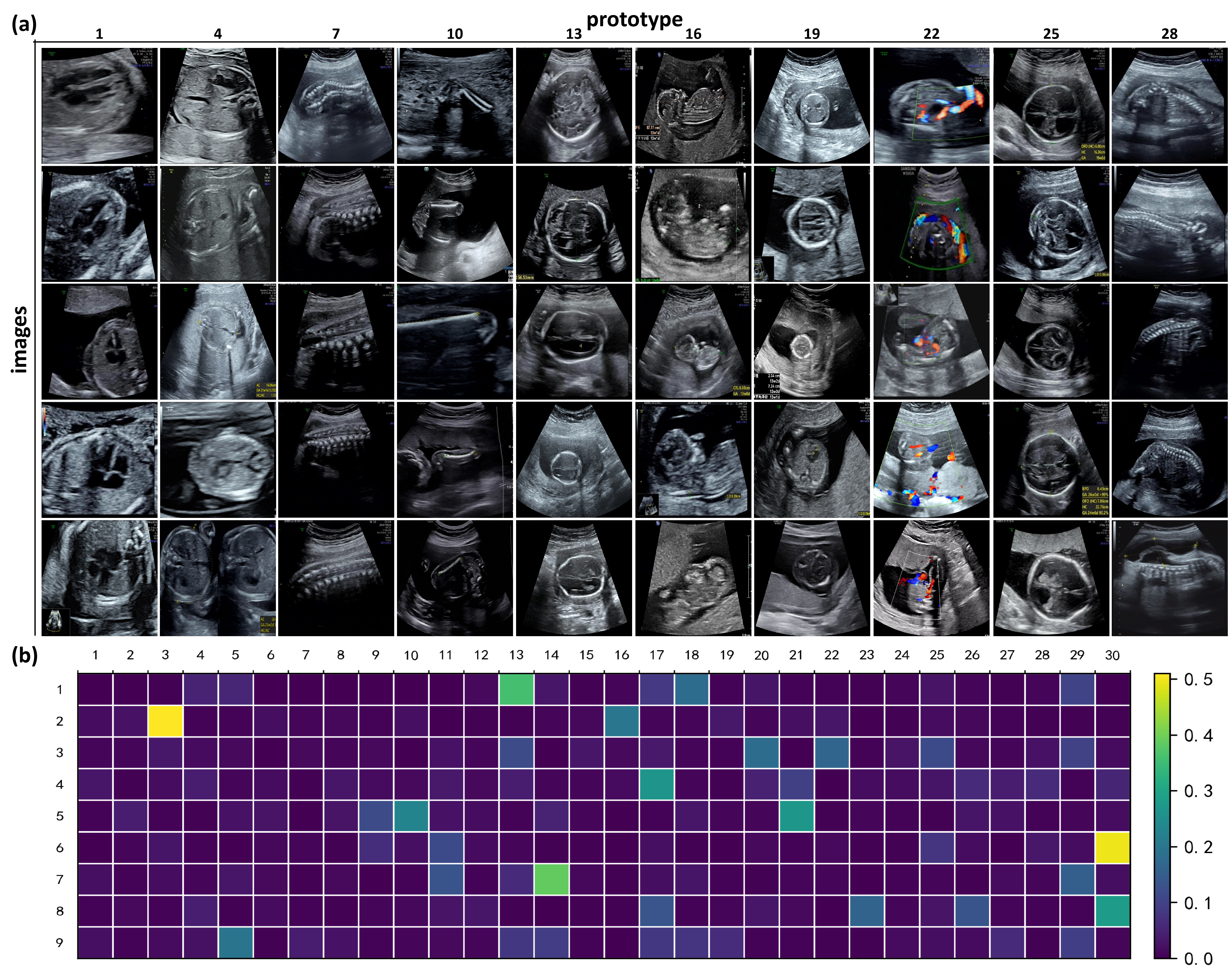}
\caption{Qualitative visualization.
(a) Test images assigned to selected plane prototypes.
(b) Class-wise relevance heatmap showing disease-plane associations.}
\label{figure3}
\end{figure}

\subsubsection{Quantitative Results}

Table~\ref{tab1} presents the quantitative comparison between AnomExpert and nine representative MIL methods on the test set.
Overall, the ViT-small variant of AnomExpert achieves the best performance across all evaluation metrics. Specifically, it achieves 86.9\% accuracy, 84.9\% precision, 83.9\% recall, 84.2\% F1-score, and 97.9\% AUC. 
Compared with the best-performing baseline, MaxNetMIL, AnomExpert improves accuracy by 1.4 percentage points (86.9\% vs. 85.5\%) and F1-score by 2.1 percentage points (84.2\% vs. 82.1\%).
Furthermore, AnomExpert demonstrates strong parameter efficiency. 
Using the ViT-tiny backbone, it achieves competitive performance (83.8\% accuracy and 97.3\% AUC) with only 5.60M parameters, fewer than all baseline models. 
These results confirm the effectiveness of AnomExpert in improving performance and parameter efficiency over existing MIL methods.

\subsubsection{Qualitative Results}

Fig.~\ref{figure3}(a) shows test images assigned to selected plane prototypes. 
Images grouped under the same prototype exhibit consistent anatomical patterns, indicating that the learned prototypes approximate anatomical plane identity without requiring plane annotations.
Fig.~\ref{figure3}(b) shows the class-wise plane selection heatmap.
The heatmap reflects the average selection probabilities of planes across all test cases.
Certain planes are selected more frequently, suggesting that they correspond to diagnostically informative anatomical planes. 
Together, these visualizations demonstrate that AnomExpert organizes images into plane representations and performs selective aggregation for prediction.

\subsubsection{Ablation Study}

\begin{table*}[t]
\centering
\caption{Component ablations of AnomExpert on the test set (mean $\pm$ std over five runs). 
All metrics are reported in percentage (\%).}
\label{tab:ablation_components}
\begin{tabular}{l|>{\centering\arraybackslash}p{1.65cm} >{\centering\arraybackslash}p{1.65cm}}
\hline
\textbf{Variant} & \textbf{Accuracy} & \textbf{F1-score} \\
\hline
Full AnomExpert & \textcolor{blue}{86.9$\pm$0.72} &  \textcolor{blue}{84.2$\pm$0.69}  \\
\textbf{Plane Prototype Learning} \\
\quad w/o prototype loss ($\lambda=0$) &84.8$\pm$1.18  &82.7$\pm$1.43  \\
\quad Softmax assignment (w/o Sinkhorn balancing) &83.4$\pm$0.72  &80.8$\pm$0.78  \\
\textbf{Disease-aware Plane Selection} \\
\quad Dense aggregation (w/o top-$k$) &86.2$\pm$1.81  &83.7$\pm$1.81  \\
\quad w/o disease bias ($B=0$) &85.1$\pm$1.22  &82.4$\pm$1.46  \\
\hline
\end{tabular}
\end{table*}

\begin{figure}[b]
\centering
\includegraphics[scale=0.7]{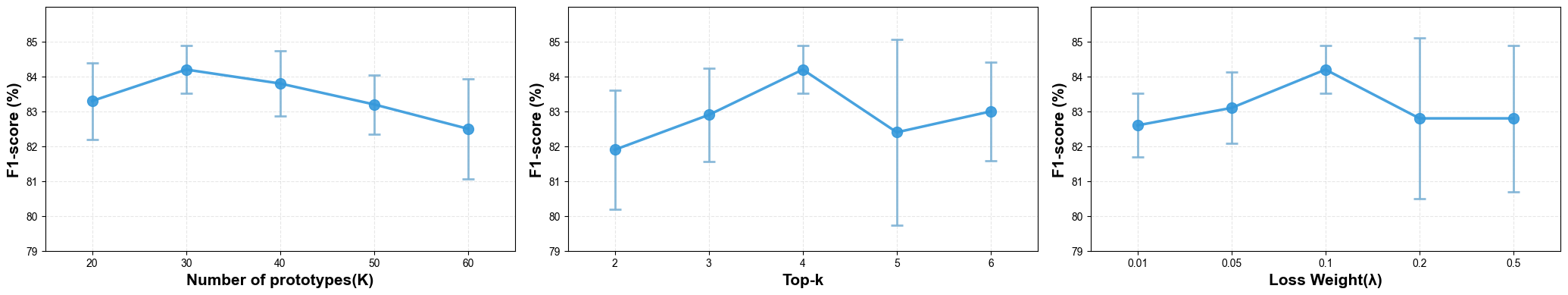}
\caption{Ablation study on key hyperparameters. F1-score versus plane prototype number $K$ (left), top-$k$ selection (middle), and loss weight $\lambda$ (right).}
\label{figure4}
\end{figure}

Table~\ref{tab:ablation_components} shows that removing prototype learning ($\lambda=0$) reduces F1-score by 1.5 points, while replacing Sinkhorn with softmax leads to a larger drop to 80.8\%. 
For disease-aware selection, dense aggregation and removing the disease bias matrix both degrade performance. 
These results indicate that both plane identification and disease-aware selection contribute to overall performance.
Fig.~\ref{figure4} illustrates the influence of key hyperparameters. 
We evaluated five candidate values for each hyperparameter, and the best performance was achieved with $K=30$, top-$k=4$, and $\lambda=0.1$.

\section{Conclusion}

We proposed AnomExpert, a prototype-driven framework for prenatal US anomaly diagnosis under case-level supervision. 
By modeling anatomical plane identification and disease-specific plane selection, the method integrates key diagnostic capabilities into MIL.
Experiments show consistent improvements over representative MIL methods, and ablation studies support the contribution of each component. 
These results highlight the importance of combining anatomical plane modeling and disease-aware plane selection for improving weakly supervised multi-plane medical image analysis.
Future research could explore enhancing the scalability of AnomExpert for more diverse datasets and integrating it with clinical workflows for real-time anomaly detection and diagnosis.

\begin{credits}
\subsubsection{\ackname}
This work is supported by the Frontier Technology Development Program of Jiangsu Province (No. BF2024078), National Natural Science Foundation of China (No. 12326619, 62572324), Science and Technology Planning Project of Guangdong Province (No. 2023A0505020002).

\subsubsection{\discintname}
The authors have no competing interests to declare that are relevant to the content of this article. 

\end{credits}

\bibliographystyle{splncs04}
\bibliography{ref}     

@inproceedings{breeze2013antenatal,
  title={Antenatal diagnosis and management of life-limiting conditions},
  author={Breeze, Andrew CG and Lees, Christoph C},
  booktitle={Seminars in Fetal and Neonatal Medicine},
  volume={18},
  number={2},
  pages={68--75},
  year={2013},
  organization={Elsevier}
}

@article{adzick2011randomized,
  title={A randomized trial of prenatal versus postnatal repair of myelomeningocele},
  author={Adzick, N Scott and Thom, Elizabeth A and Spong, Catherine Y and Brock III, John W and Burrows, Pamela K and Johnson, Mark P and Howell, Lori J and Farrell, Jody A and Dabrowiak, Mary E and Sutton, Leslie N and others},
  journal={New England Journal of Medicine},
  volume={364},
  number={11},
  pages={993--1004},
  year={2011},
  publisher={Mass Medical Soc}
}

@article{ohye2016current,
  title={Current therapy for hypoplastic left heart syndrome and related single ventricle lesions},
  author={Ohye, Richard G and Schranz, Dietmar and D’Udekem, Yves},
  journal={Circulation},
  volume={134},
  number={17},
  pages={1265--1279},
  year={2016},
  publisher={Lippincott Williams \& Wilkins Hagerstown, MD}
}

@article{cook2008prenatal,
  title={Prenatal management of anencephaly},
  author={Cook, Rebecca J and Erdman, Joanna N and Hevia, Martin and Dickens, Bernard M},
  journal={International Journal of Gynecology \& Obstetrics},
  volume={102},
  number={3},
  pages={304--308},
  year={2008},
  publisher={Elsevier}
}

@article{salomon2022isuog,
  title={ISUOG Practice Guidelines (updated): performance of the routine mid-trimester fetal ultrasound scan},
  author={Salomon, LJ and Alfirevic, Z and Berghella, V and Bilardo, CM and Chalouhi, GE and Costa, F Da Silva and Hernandez-Andrade, E and Malinger, G and Munoz, H and Paladini, D and others},
  journal={Ultrasound in Obstetrics and Gynecology},
  volume={59},
  number={6},
  pages={840--856},
  year={2022},
  publisher={Wiley}
}

@article{sarris2012intra,
  title={Intra-and interobserver variability in fetal ultrasound measurements},
  author={Sarris, I and Ioannou, C and Chamberlain, P and Ohuma, E and Roseman, F and Hoch, L and Altman, DG and Papageorghiou, AT and International Fetal and Newborn Growth Consortium for the 21st Century (INTERGROWTH-21st)},
  journal={Ultrasound in obstetrics \& gynecology},
  volume={39},
  number={3},
  pages={266--273},
  year={2012},
  publisher={Wiley Online Library}
}

@article{guo2022fetal,
  title={Fetal ultrasound standard plane detection with coarse-to-fine multi-task learning},
  author={Guo, Juncheng and Tan, Guanghua and Wu, Fan and Wen, Huaxuan and Li, Kenli},
  journal={IEEE Journal of Biomedical and Health Informatics},
  volume={27},
  number={10},
  pages={5023--5031},
  year={2022},
  publisher={IEEE}
}

@article{arnaout2021ensemble,
  title={An ensemble of neural networks provides expert-level prenatal detection of complex congenital heart disease},
  author={Arnaout, Rima and Curran, Lara and Zhao, Yili and Levine, Jami C and Chinn, Erin and Moon-Grady, Anita J},
  journal={Nature medicine},
  volume={27},
  number={5},
  pages={882--891},
  year={2021},
  publisher={Nature Publishing Group US New York}
}

@article{coronado2023automatic,
  title={Automatic deep learning-based pipeline for automatic delineation and measurement of fetal brain structures in routine mid-trimester ultrasound images},
  author={Coronado-Gutierrez, David and Eixarch, Elisenda and Monterde, Elena and Matas, Isabel and Traversi, Paola and Gratacos, Eduard and Bonet-Carne, Elisenda and Burgos-Artizzu, Xavier P},
  journal={Fetal diagnosis and therapy},
  volume={50},
  number={6},
  pages={480--490},
  year={2023},
  publisher={S. Karger AG}
}

@article{barbosa2024multiple,
  title={Multiple instance learning in medical images: a systematic review},
  author={Barbosa, Dalila and Ferreira, Marcos and Junior, Geraldo Braz and Salgado, Marta and Cunha, Ant{\'o}nio},
  journal={IEEE Access},
  volume={12},
  pages={78409--78422},
  year={2024},
  publisher={IEEE}
}

@inproceedings{huang2023detecting,
  title={Detecting heart disease from multi-view ultrasound images via supervised attention multiple instance learning},
  author={Huang, Zhe and Wessler, Benjamin S and Hughes, Michael C},
  booktitle={Machine Learning for Healthcare Conference},
  pages={285--307},
  year={2023},
  organization={PMLR}
}

@inproceedings{ilse2018attention,
  title={Attention-based deep multiple instance learning},
  author={Ilse, Maximilian and Tomczak, Jakub and Welling, Max},
  booktitle={International conference on machine learning},
  pages={2127--2136},
  year={2018},
  organization={PMLR}
}

@incollection{ilse2020deep,
  title={Deep multiple instance learning for digital histopathology},
  author={Ilse, Maximilian and Tomczak, Jakub M and Welling, Max},
  booktitle={Handbook of Medical Image Computing and Computer Assisted Intervention},
  pages={521--546},
  year={2020},
  publisher={Elsevier}
}

@article{shao2021transmil,
  title={Transmil: Transformer based correlated multiple instance learning for whole slide image classification},
  author={Shao, Zhuchen and Bian, Hao and Chen, Yang and Wang, Yifeng and Zhang, Jian and Ji, Xiangyang and others},
  journal={Advances in neural information processing systems},
  volume={34},
  pages={2136--2147},
  year={2021}
}

@inproceedings{li2021dual,
  title={Dual-stream multiple instance learning network for whole slide image classification with self-supervised contrastive learning},
  author={Li, Bin and Li, Yin and Eliceiri, Kevin W},
  booktitle={Proceedings of the IEEE/CVF conference on computer vision and pattern recognition},
  pages={14318--14328},
  year={2021}
}

@article{javed2022additive,
  title={Additive mil: Intrinsically interpretable multiple instance learning for pathology},
  author={Javed, Syed Ashar and Juyal, Dinkar and Padigela, Harshith and Taylor-Weiner, Amaro and Yu, Limin and Prakash, Aaditya},
  journal={Advances in Neural Information Processing Systems},
  volume={35},
  pages={20689--20702},
  year={2022}
}

@article{zhang2024mamba2mil,
  title={Mamba2mil: State space duality based multiple instance learning for computational pathology},
  author={Zhang, Yuqi and Zhang, Xiaoqian and Wang, Jiakai and Yang, Yuancheng and Peng, Taiying and Tong, Chao},
  journal={arXiv preprint arXiv:2408.15032},
  year={2024}
}

@inproceedings{shiku2025ordinal,
  title={Ordinal Multiple-instance Learning for Ulcerative Colitis Severity Estimation with Selective Aggregated Transformer},
  author={Shiku, Kaito and Nishimura, Kazuya and Suehiro, Daiki and Tanaka, Kiyohito and Bise, Ryoma},
  booktitle={2025 IEEE/CVF Winter Conference on Applications of Computer Vision (WACV)},
  pages={4290--4299},
  year={2025},
  organization={IEEE}
}

@inproceedings{liang2025medical,
  title={Medical-Knowledge Driven Multiple Instance Learning for Classifying Severe Abdominal Anomalies on Prenatal Ultrasound},
  author={Liang, Huanwen and Xu, Jingxian and Zhang, Yuanji and Huang, Yuhao and Zhang, Yuhan and Yang, Xin and Li, Ran and Deng, Xuedong and Liu, Yanjun and Tao, Guowei and others},
  booktitle={International Conference on Medical Image Computing and Computer-Assisted Intervention},
  pages={344--354},
  year={2025},
  organization={Springer}
}

@article{cuturi2013sinkhorn,
  title={Sinkhorn distances: Lightspeed computation of optimal transport},
  author={Cuturi, Marco},
  journal={Advances in neural information processing systems},
  volume={26},
  year={2013}
}

\end{document}